\documentclass{article}

\PassOptionsToPackage{numbers,compress}{natbib}
\IfFileExists{neurips_2024.sty}{
    \usepackage[preprint]{neurips_2024}%
}{
    \usepackage[margin=1in]{geometry}
    \usepackage[numbers,compress]{natbib}
    \usepackage{microtype}
}

\usepackage[utf8]{inputenc}
\usepackage[T1]{fontenc}
\usepackage{amsmath,amssymb,amsfonts}
\usepackage{algorithmic}
\usepackage{graphicx}
\usepackage{textcomp}
\usepackage{xcolor}
\usepackage{tikz}
\usepackage{subcaption}
\usepackage{hyperref}
\usepackage{url}
\usetikzlibrary{shapes, shapes.geometric, arrows.meta, positioning, chains, fit, calc}
\begin{document}

	\title{Exploring Expert Specialization through Unsupervised Training in Sparse Mixture of Experts}

\author{
    Strahinja Nikolic\textsuperscript{1},
    Ilker Oguz\textsuperscript{2},
    and Demetri Psaltis\textsuperscript{3} \\
    \textsuperscript{1}School of Engineering (STI),
    \textsuperscript{2}Laboratory of Applied Photonics Devices (LAPD),
    \textsuperscript{3}Optics Laboratory \\
    École Polytechnique Fédérale de Lausanne (EPFL) \\
    \texttt{\{strahinja.nikolic, ilker.oguz, demetri.psaltis\}@epfl.ch}
}

\maketitle

\begin{abstract}
Understanding the internal organization of neural networks remains a fundamental challenge in deep learning interpretability. We address this challenge by exploring a novel Sparse Mixture of Experts Variational Autoencoder (SMoE-VAE) architecture. We test our model on the QuickDraw dataset, comparing unsupervised expert routing against a supervised baseline guided by ground-truth labels. Surprisingly, we find that unsupervised routing consistently achieves superior reconstruction performance. The experts learn to identify meaningful sub-categorical structures that often transcend human-defined class boundaries. Through t-SNE visualizations and reconstruction analysis, we investigate how MoE models uncover fundamental data structures that are more aligned with the model's objective than predefined labels. Furthermore, our study on the impact of dataset size provides insights into the trade-offs between data quantity and expert specialization, offering guidance for designing efficient MoE architectures.
\end{abstract}

\section{Introduction}

Mixture of Experts (MoE) architectures, first introduced by Jacobs et al.~\cite{jacobs1991adaptive}, decompose computation across specialized sub-networks, or ``experts.'' This paradigm has recently achieved remarkable success in scaling deep learning models to unprecedented sizes, enabling state-of-the-art performance in areas like large language modeling~\cite{fedus2022switch, lepikhin2020gshard, survey_moe_llm}. The promise of MoE models extends beyond computational efficiency: the explicit routing of inputs to different experts can offer a natural window to help us understand how neural networks  organize and process data.

However, as MoE models become more powerful and complex, understanding what each expert has learned or how routing decisions relate to meaningful data structure remains a fundamental challenge~\cite{survey_moe_llm, understanding_moe, benefits_routing_moe}. This opacity is particularly concerning as these systems are increasingly deployed in critical applications where understanding model behavior is essential. This gap between performance and interpretability points to  the need for new methods to analyze the internal mechanisms of MoE models.

Recent theoretical advances have begun to illuminate the fundamental mechanisms underlying expert specialization. The work by \cite{understanding_moe} demonstrates that MoE architectures possess an inherent capacity to train  experts specializing in different data clusters when trained with gradient descent. Their analysis reveals that this specialization emerges automatically from the optimization dynamics, suggesting that expert assignment patterns are likely to  reflect fundamental properties of the data distribution.

Yet, such analysis remain sparse in the literature. For instance, a recent comprehensive survey~\cite{survey_moe_llm} argues that to ensure MoE models are transparent and trustworthy, the field urgently needs new methods to visualize and explain what individual experts learn and how they interact. Improving this aspect of interpretability is considered critical for bolstering our understanding and advancing the responsible development of MoE architectures.

To address this challenge, we combine Sparse Mixture of Experts with Variational Autoencoders (VAEs), creating an architecture tailored for the direct analysis of expert specialization. Our approach is partially inspired by MoE-Sim-VAE~\cite{moe_sim_vae}, which demonstrated the potential of MoE-VAEs for clustering. However, we repurpose this architecture with a distinct goal: to analyze the phenomenon of expert specialization itself and the mechanisms that drive it. The architecture's inherent ability to discover underlying data structures makes it an ideal tool for this investigation.

Our investigation reveals a surprising finding: when experts are allowed to specialize according to the natural structure present in data (unsupervised routing), they form clusters that are more effective for the model's objective than when grouped by human-provided labels (supervised routing). As we will demonstrate, these data-driven clusters are closer to being linearly separable and result in superior reconstruction performance. This suggests that expert specialization in MoE architectures can serve as a powerful tool for data structure discovery, uncovering organizational principles that differ from conventional categorizations.

\textbf{Contributions.}  We make the following key contributions:

\begin{itemize}
    \item We introduce SMoE-VAE, a sparse Mixture-of-Experts Variational Autoencoder tailored for interpretability and analysis of expert specialization. The architecture uses a shared high-capacity encoder with lightweight decoder experts, a MLP as a gating unit which receives latent space vectors as input, and is trained with entropy and batch-level load balancing loss.
    \item To the best of our knowledge, we present the first controlled comparison between supervised (ground-truth-label–guided) and unsupervised routing. Unsupervised routing consistently achieves lower reconstruction loss and discovers clusters in the latent space that are closer to being linearly  separable compared to supervised training with a labeled database. 
    \item We provide an analysis framework that explains \emph{why} unsupervised training  prevails: (i) latent-space visualizations (t-SNE) contrasting expert assignments and class labels, (ii) linear probes that quantify separability of expert vs. class partitions, and (iii) qualitative visualization of reconstructions revealing cross-class and intra-class specializations.
    \item We study the joint effect of the number of experts and the number of samples per expert on performance. We show that gains cannot be attributed to data quantity alone; the structure of the data each expert sees and the resulting degree of specialization are critical, with over-fragmentation and under-fragmentation both leading to reduced performance.
\end{itemize}

\begin{figure}[htbp]
    \centering
    \resizebox{\textwidth}{!}{\begin{tikzpicture} [
        auto,
        node distance=1.0cm and 0.8cm,
        base/.style={draw, rectangle, minimum height=1.0cm, minimum width=1.5cm, align=center, font=\small},
        encoder/.style={base, fill=blue!20},
        decoder/.style={base, fill=green!20},
        gate/.style={base, fill=orange!20, rounded corners},
        latent/.style={draw, circle, fill=yellow!30, minimum size=0.8cm, font=\small},
        data/.style={draw, rectangle, fill=gray!10, minimum height=0.6cm, minimum width=0.8cm, font=\small},
        sum/.style={draw, circle, minimum size=0.5cm},
        arrow/.style={-Stealth, thick},
        dashedarrow/.style={-Stealth, thick, dashed}
    ]

    \node[data] (input) {$x$};
    \node[encoder, right=of input] (enc) {Shared\\Encoder};
    \node[latent, right=of enc] (z) {$z$};
    
    \node[gate, right=of z, xshift=0.4cm] (gate) {Gating Network\\(MLP)};
    
    \node[decoder, right=of gate, yshift=1.5cm, xshift=0.4cm] (dec1) {Decoder 1};
    \node[decoder, right=of gate, yshift=0.0cm, xshift=0.4cm] (dec2) {Decoder 2};
    \node[font=\large, below=0.15cm of dec2] (dots1) {$\vdots$};
    \node[decoder, below=0.15cm of dots1] (dece) {Decoder e};
    \node[font=\large, below=0.15cm of dece] (dots2) {$\vdots$};
    \node[decoder, below=0.15cm of dots2] (decE) {Decoder E};
    
    \node[data, right=of dec1] (out1) {$\hat{x}_1$};
    \node[data, right=of dec2] (out2) {$\hat{x}_2$};
    \node[data, right=of dece] (oute) {$\hat{x}_e$};
    \node[data, right=of decE] (outE) {$\hat{x}_E$};

    \node[sum, right=of out2, xshift=0.3cm] (summation) {$\sum$};
    
    \node[data, right=of summation, xshift=0.3cm] (output) {$\hat{x}$};

    \node[draw, dashed, inner sep=0.2cm, label={[yshift=0.1cm]above:Experts}, fit=(dec1) (decE)] (experts_box) {};

    \draw[arrow] (input) -- (enc);
    \draw[arrow] (enc) -- (z);
    
    \draw[arrow] (z) -- (gate);
    
    \draw[arrow] (gate) -- node[below left, font=\small] {$p_e$} (dece);
    \draw[dashedarrow] (gate) -- node[above left, font=\small] {$p_1$} (dec1);
    \draw[dashedarrow] (gate) -- node[above left, font=\small] {$p_2$} (dec2);
    \draw[dashedarrow] (gate) -- node[below left, font=\small] {$p_E$} (decE);

    \draw[arrow] (dec1) -- (out1);
    \draw[arrow] (dec2) -- (out2);
    \draw[arrow] (dece) -- (oute);
    \draw[arrow] (decE) -- (outE);
    
    \draw[dashedarrow] (out1) -- (summation);
    \draw[dashedarrow] (out2) -- (summation);
    \draw[arrow] (oute) -- (summation);
    \draw[dashedarrow] (outE) -- (summation);
    
    \draw[arrow] (summation) -- (output);

    \end{tikzpicture}
    }
    \caption{The SMoE-VAE architecture. A shared encoder maps the input $x$ to a latent representation $z$. The gating network receives $z$ and routes it to specialized decoders. During training all decoders are activated and the final output is the sum of all outputs weighted by gating network probabilities, while during inference only one expert is activated to produce $\hat{x}$.}
    \label{fig:smoe_vae_arch}
\end{figure}
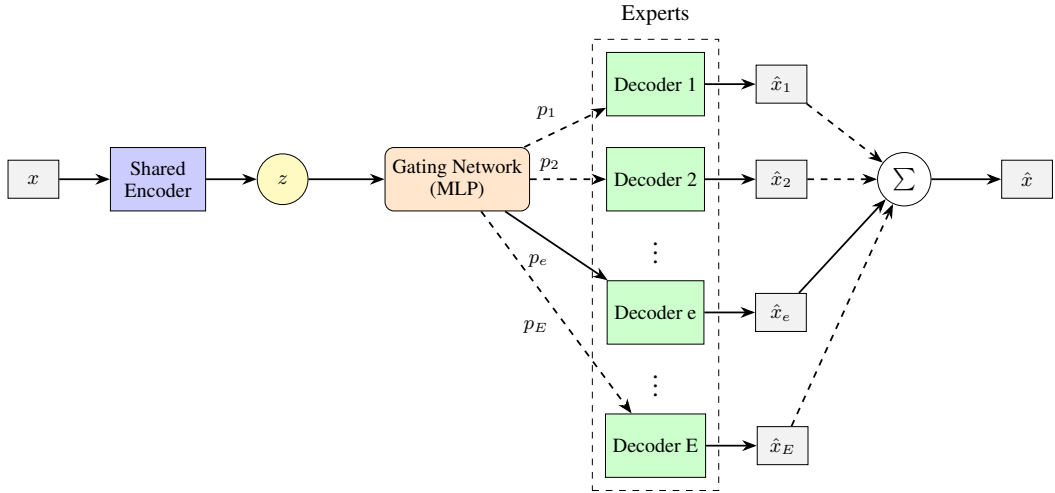

\section{Related Work}

\subsection{Mixture of Experts and Interpretability}

Mixture of Experts (MoE) architectures have emerged as a powerful paradigm for scaling neural networks while maintaining computational efficiency. Originally introduced by Jacobs et al.~\cite{jacobs1991adaptive}, MoE models partition the input space among specialized experts, with a gating network determining the routing of inputs to appropriate experts. Recent advances have demonstrated the effectiveness of sparse MoE in large language models~\cite{fedus2022switch, lepikhin2020gshard}, where only a subset of experts are activated for each input.

However, as highlighted in recent surveys, the interpretability of MoE models remains a significant challenge. The inherent complexity of MoE models, coupled with their dynamic gating of inputs to specialized experts, poses substantial obstacles to understanding their decision-making processes~\cite{survey_moe_llm}. This interpretability gap becomes particularly problematic in applications where comprehending the rationale behind model decisions is essential. Current approaches to MoE interpretability focus primarily on analyzing expert utilization patterns and load balancing, but provide limited insight into what each expert has actually learned to specialize in.

From a theoretical perspective, recent work by~\cite{understanding_moe} provides crucial insights into the mechanisms of expert specialization. They address why experts diversify rather than collapse into a single unit and how the router learns to dispatch data. Their work suggests that experts specialize to exploit underlying cluster structures in the data, and the router learns to identify cluster-specific features. However, their analysis relies on synthetic datasets with linearly separable clusters and uses a linear gating mechanism. This leaves open questions about specialization in more complex, high-dimensional settings with non-linear structures, which our work explores.

One paper that touches on this topic is work from C.Riquelme et al. ~\cite{riquelme2021scaling} where they used sparsely gated MoE for Computer Vision. By having access to a huge dataset ($\sim$305M) of labeled images they were able to show that deeper routing decisions correlate better with image classes. This indicates that even in large networks, with multiple MoE layers, the routers are able to partition the input space in a way that correlates well with human labels.

An important question in MoE design concerns the choice of routing strategy. While prior work has shown that learned routing outperforms fixed strategies~\cite{benefits_routing_moe}, a deeper understanding of the resulting specializations is needed. The availability of labeled data provides a unique opportunity to analyze these unsupervised, data-driven clusters by comparing them to a supervised baseline where routing is determined by ground-truth categories. Such a comparison, serves as a powerful analytical tool. It allows us to rigorously evaluate the clusters discovered by unsupervised methods against human-defined categories, offering  insights into why and how unsupervised expert specialization can be so effective.

\subsection{MoE combined with Variational Autoencoders for Data Structure Analysis}

Variational Autoencoders~\cite{kingma2014auto} provide an ideal testbed for our investigation, as their explicit latent space representation and generative capabilities enable direct visualization of how experts partition and specialize on different aspects of the data distribution. Kopf A. et al. introduced MoE-Sim-VAE~\cite{moe_sim_vae} which has demonstrated the potential of combining MoE architectures with VAEs for clustering purposes. Their approach places multiple decoder experts after the latent bottleneck and encourages expert specialization through similarity-based constraints. This work establishes that expert specialization in generative models can serve as an effective clustering mechanism. Their approach requires external guidance, such as pre-computed similarity matrices or explicit clustering objectives, to achieve meaningful expert assignment.

One  approach related to ours  is MIXAE~\cite{zhang2017moe}, a framework that uses a mixture of autoencoders for unsupervised clustering. In contrast to our architecture, where only the decoders are experts, MIXAE treats entire autoencoders as experts. This design increases the complexity of the gating unit, which must process concatenated latent codes from all autoencoders. Similar to MoE-Sim-VAE, the primary focus of MIXAE is on clustering performance rather than on the interpretability of the MoE architecture itself, leaving the inner workings of expert specialization largely unexplored.

Another significant contribution, which differs from the previous two mentioned, is the Variational Mixture-of-Experts Autoencoders (MMVAE)~\cite{shi2019variationalmixtureofexpertsautoencoders}, which explores multi-modal deep generative models using MoE architectures. The MMVAE approach demonstrates how expert specialization can emerge across different data modalities, providing insights into how MoE architectures can naturally partition complex, multi-modal data distributions. While their focus is on cross-modal generation and representation learning, their work establishes important principles about how expert assignment can reveal underlying data structure.

Our work is motivated by the hypothesis that when properly designed, SMoE-VAE architectures can serve as powerful tools for data structure analysis, discovering clusters and specializations that may be more fundamental to the data distribution than human-imposed labels. Rather than optimizing for generation quality alone, we focus on understanding what experts learn when given the freedom to specialize according to the structure present in the data. The visual nature of image data makes this approach particularly interesting, as expert specializations can be directly assessed through reconstructed images, enabling immediate feedback about what each expert has learned to generate.

\section{Method}

\subsection{SMoE-VAE Architecture}

Our approach combines Variational Autoencoders with Sparse Mixture of Experts to enable interpretable analysis of expert specialization patterns. The architecture consists of three components: a shared convolutional encoder, a gating network, and multiple decoder experts that specialize on different aspects of the data distribution.

\subsubsection{Encoder-Decoder Design}

A key architectural feature is the asymmetric capacity allocation between encoder and decoder components. We employ a single, high-capacity convolutional encoder to process input images, ensuring effective feature extraction and meaningful latent space clustering. This design choice is crucial since all expert routing decisions depend on the quality of latent representations produced by the shared encoder.

In contrast, we use smaller, specialized decoder experts that focus on reconstructing specific data patterns. This asymmetry serves two purposes: (1) it concentrates representational capacity where it is most needed for clustering, and (2) it encourages experts to specialize on distinct reconstruction tasks rather than learning broad representations.

\subsubsection{Gating Network}

The gating network operates on latent representations $z \in \mathbb{R}^d$ to produce expert selection probabilities. It consists of a three-layer fully connected multilayer perceptron:

\begin{align*}
h_1 &= \text{ReLU}(W_1 z + b_1) \\
h_2 &= \text{ReLU}(W_2 h_1 + b_2) \\
\text{logits} &= W_3 h_2 + b_3
\end{align*}

where $W_1 \in \mathbb{R}^{64 \times d}$, $W_2 \in \mathbb{R}^{32 \times 64}$, and $W_3 \in \mathbb{R}^{E \times 32}$ with $E$ being the number of experts. Note that, since the gating network operates on the latent space the additional computational cost of this  network is minimal compared to the rest of the network.

\subsubsection{Soft vs Hard Gating Strategy}

To bridge the gap between training and inference, we employ a dual gating strategy. During training, we use soft gating where expert outputs are weighted by softmax probabilities:

\begin{align*}
p_e &= \frac{\exp(\text{logits}_e)}{\sum_{i=1}^E \exp(\text{logits}_i)} \\
\hat{x} &= \sum_{e=1}^E p_e \cdot \text{Decoder}_e(z)
\end{align*}

During inference, we switch to hard gating for efficiency and interpretability:

\begin{align*}
e^* &= \arg\max_e \text{logits}_e \\
\hat{x} &= \text{Decoder}_{e^*}(z)
\end{align*}

This approach ensures that the model learns to make confident expert selections while maintaining differentiability during training.

\subsection{Loss Function Design}

Our loss function combines the standard VAE objective~\cite{kingma2014auto} with novel regularization terms that encourage both load balancing and decisive expert selection:

\begin{align*}
\mathcal{L}_{\text{total}} &= \mathcal{L}_{\text{recon}} + \beta \mathcal{L}_{\text{KL}} + \alpha \mathcal{L}_{\text{gating}}
\end{align*}

where $\mathcal{L}_{\text{recon}}$ is the Mean Squared Error (MSE) reconstruction loss, $\mathcal{L}_{\text{KL}}$ is the KL divergence regularization~\cite{kingma2014auto}, and $\mathcal{L}_{\text{gating}}$ combines load balancing and entropy regularization.

\subsubsection{Load Balancing Loss}

To prevent expert collapse and ensure diverse specialization, we introduce a load balancing term that encourages uniform expert utilization across batches:

\begin{align*}
\bar{p}_e &= \frac{1}{N} \sum_{n=1}^N p_{n,e} \\
\mathcal{L}_{\text{balance}} &= E \cdot \text{MSE}(\bar{p}, \mathbf{u})
\end{align*}

where $\bar{p} = [\bar{p}_1, \ldots, \bar{p}_E]$ represents average expert probabilities across the batch, $p_n$ are softmax probabilities for sample with index $n$, $\mathbf{u} = [1/E, \ldots, 1/E]$ is the uniform distribution, and $N$ is the batch size. The scaling factor $E$ ensures consistent regularization strength across different numbers of experts.

\subsubsection{Entropy Regularization}

To encourage sharp expert selections and minimize the train-inference gap, we minimize the entropy of per-sample expert distributions:

\begin{align*}
\mathcal{L}_{\text{entropy}} &= \frac{1}{N} \sum_{n=1}^N H(p_n) \\
H(p_n) &= -\sum_{e=1}^E p_{n,e} \log(p_{n,e} + \epsilon)
\end{align*}

where $\epsilon = 10^{-8}$ provides numerical stability. Minimizing entropy encourages the model to produce near-Dirac delta distributions, where one expert receives most of the probability mass. This ensures that soft gating during training closely approximates hard gating during inference.

The combined gating loss is:
\begin{align*}
\mathcal{L}_{\text{gating}} = \lambda_{\text{balance}} \mathcal{L}_{\text{balance}} + \lambda_{\text{entropy}} \mathcal{L}_{\text{entropy}}
\end{align*}

\section{Experimental Setup}

\subsection{Choice of Dataset and Preprocessing}

We conduct our experiments on the QuickDraw dataset~\cite{quickdraw}, a large-scale collection of hand-drawn sketches created by users worldwide. QuickDraw is particularly well-suited for our interpretability analysis for several key reasons. First, it provides abundant data with tens of thousands of samples per category, enabling robust training of expert networks and reliable statistical analysis. Second, the availability of ground truth category labels allows for direct comparison between supervised and unsupervised expert routing approaches. Third, and most importantly for our research, QuickDraw contains natural imperfections and variations that enable meaningful sub-clustering within categories: sketches sometimes exhibit ambiguous category membership (e.g., simplified cat faces that resemble generic faces), and the same object can be drawn in multiple styles (e.g., cats drawn as face-only sketches versus full-body representations). These characteristics allow unsupervised expert routing to discover subclusters that can be more reconstruction-relevant than rigid categorical boundaries.

From the full dataset, we select five diverse categories that exhibit varying levels of visual complexity and inter-category similarity: \textit{face}, \textit{eye}, \textit{cat}, \textit{snowflake}, and \textit{pencil}. This selection provides a balanced mix of organic shapes (faces, eyes, cats), geometric patterns (snowflakes), and linear objects (pencils), enabling comprehensive analysis of expert specialization across different visual structures.

All sketches are preprocessed by converting vector drawings to 28×28 grayscale images, ensuring computational efficiency while preserving essential visual features. Unless otherwise specified, we use 70,000 samples per category, providing a dataset of 350,000 total images for training and evaluation.  In order to evaluate the impact of the  database, we systematically vary the number of samples from 5\% to 100\% of the full dataset while maintaining balanced representation across all categories.

\subsection{Model Configuration and Training Protocol}

Our implementation uses PyTorch~\cite{pytorch} and employs convolutional neural networks for both encoder and decoder components. The shared encoder processes 28×28 input images and maps them to a 32-dimensional latent space, while individual decoder experts reconstruct images from the latent representations.

Training is conducted for 20 epochs across all experiments to ensure fair comparison while avoiding overfitting. We use the Adam optimizer~\cite{adam2014} with a learning rate of $10^{-4}$. The loss function hyperparameters are set as follows: $\beta = 0.1$ for KL divergence regularization, $\lambda_{\text{balance}} = 200$ for load balancing, and $\lambda_{\text{entropy}} = 400$ for entropy regularization. These values were chosen to balance reconstruction quality with effective expert specialization and load balancing.

For supervised baseline comparisons, we train the gating network to route samples based on ground truth category labels rather than learned latent representations, while maintaining identical model capacity and training procedures.

For dataset size impact studies, we report optimal expert counts and corresponding reconstruction losses across different data regimes. All results are averaged over multiple random seeds to ensure statistical reliability, with error bars representing standard deviation across runs.

\section{Results and Analysis}

\subsection{Unsupervised vs Supervised Expert Routing}

\begin{figure}[htbp]
\centering
\includegraphics[width=0.9\linewidth]{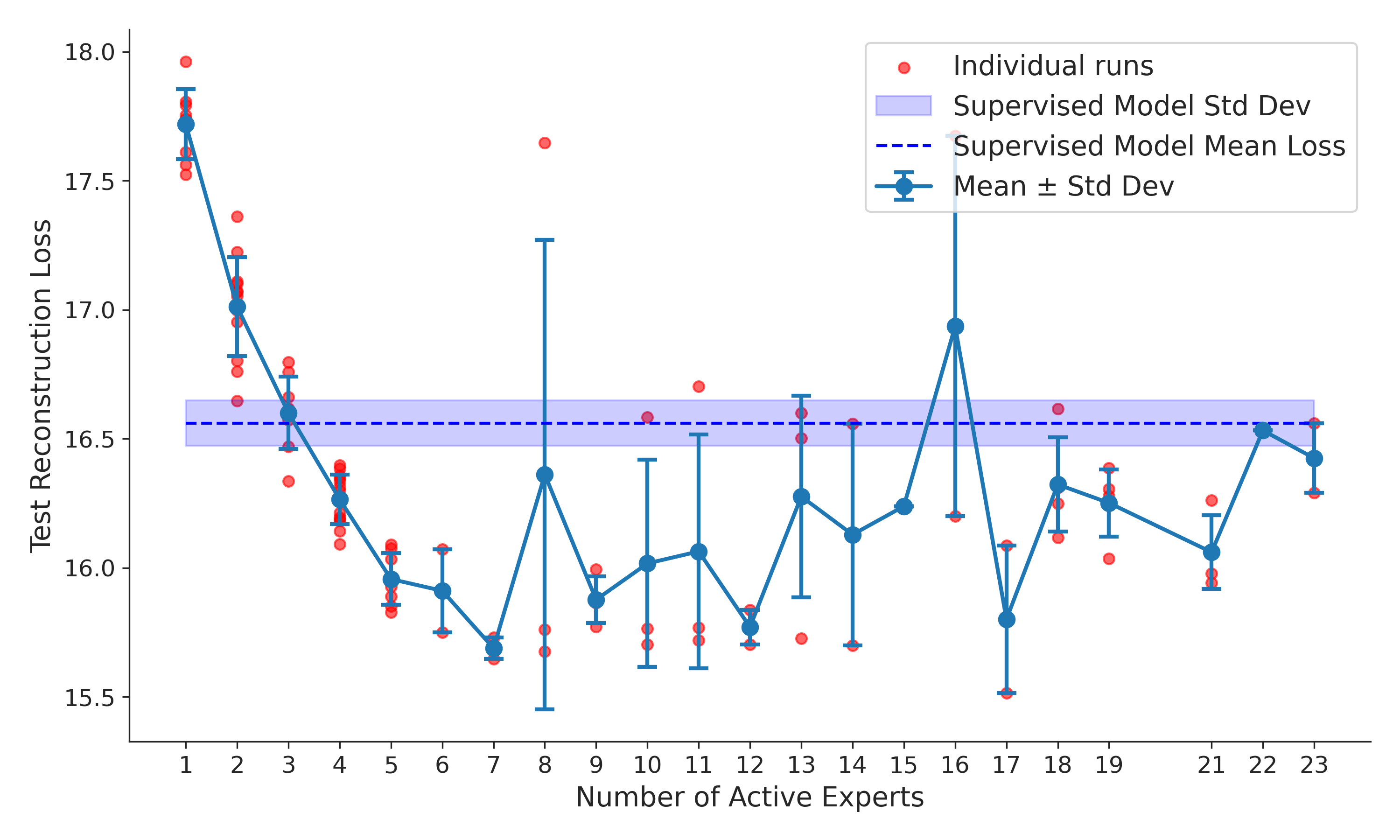}
\caption{Test reconstruction loss: unsupervised expert routing (blue) vs. supervised routing based on ground truth labels (purple dashed). Unsupervised peaks around 7 experts and outperforms the supervised baseline constrained to 5 experts.}
\label{fig:supervised_vs_unsupervised}
\end{figure}

Our primary finding demonstrates that unsupervised expert routing consistently outperforms supervised routing based on human-provided labels. Figure~\ref{fig:supervised_vs_unsupervised} presents test reconstruction loss as a function of the number of active experts, comparing our unsupervised SMoE-VAE approach (blue line with error bars) against a supervised baseline where expert assignment is determined by ground truth class labels.

The results reveal several key insights. First, the unsupervised approach achieves significantly lower reconstruction loss across most expert configurations, with the optimal performance around 7 experts reaching a MSE loss on test set near 15.7, better than the supervised approach's 16.6. The supervised model, constrained to exactly 5 experts corresponding to the 5 QuickDraw categories, represents the standard approach of aligning experts with human-defined labels.

Second, the unsupervised approach exhibits a clear performance trend: reconstruction loss decreases substantially from 1 to approximately 7 experts, then gradually degrades beyond this optimal range, suggesting that too many experts lead to over-fragmentation of the data representation. Notably, the optimal number of experts (7) differs from the number of ground truth categories (5), providing evidence that the model discovers a more nuanced data organization than human categorical labels.

The performance variance (shown by error bars) demonstrates that both approaches achieve consistent results across multiple training runs, but the unsupervised method consistently outperforms the supervised baseline.

These results support our hypothesis that unsupervised expert specialization can discover data structure that is fundamental to the underlying distribution that is not present in the human-imposed categorizations, validating the theoretical predictions from~\cite{understanding_moe} about MoE's natural cluster discovery capabilities in real-world image data.

\subsection{Expert Specialization Analysis}

\begin{figure}[htbp]
\centering
\begin{subfigure}[b]{0.48\linewidth}
    \centering
    \includegraphics[width=\linewidth]{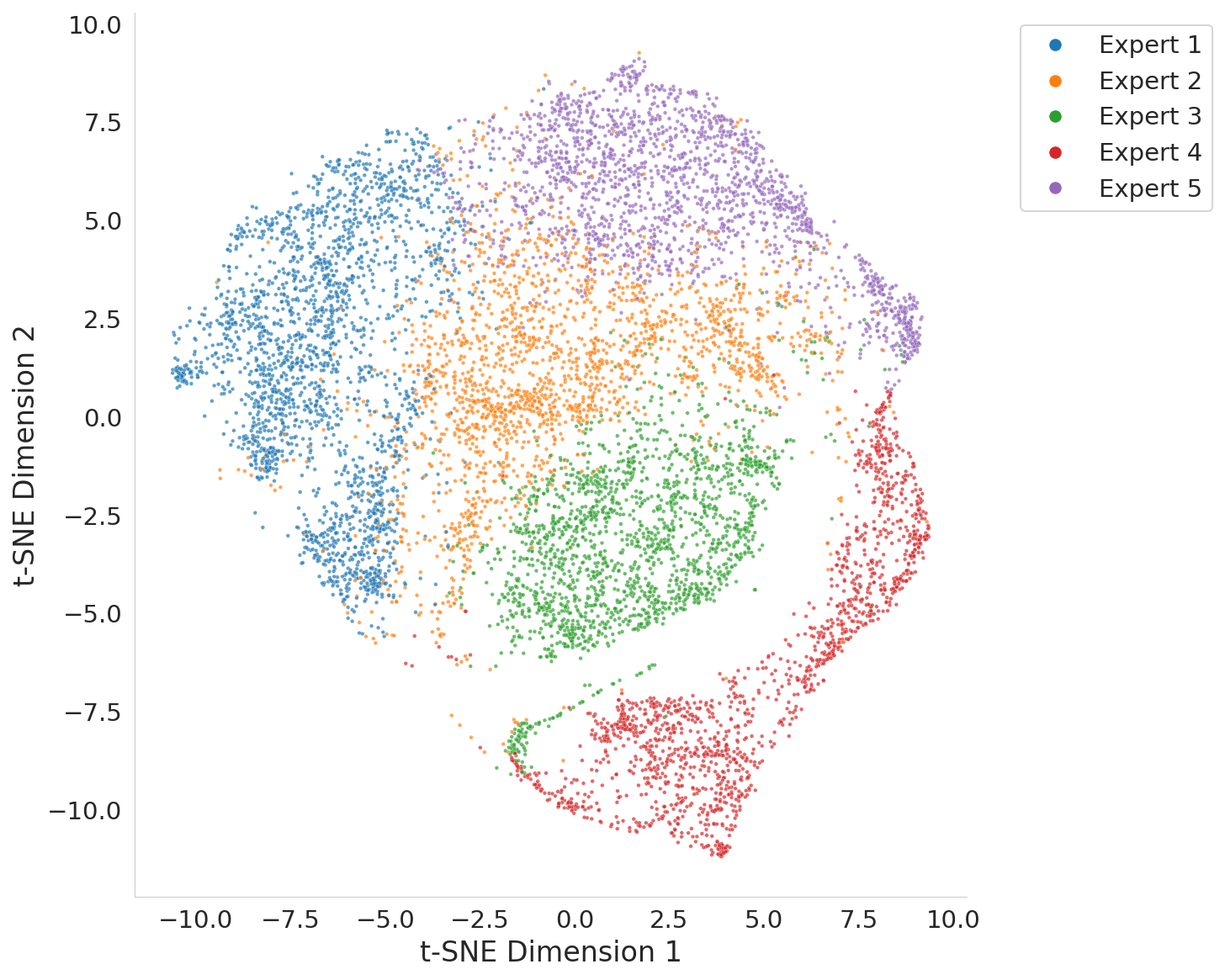}
    \caption{Colored by Expert Assignment}
\end{subfigure}\hfill
\begin{subfigure}[b]{0.48\linewidth}
    \centering
    \includegraphics[width=\linewidth]{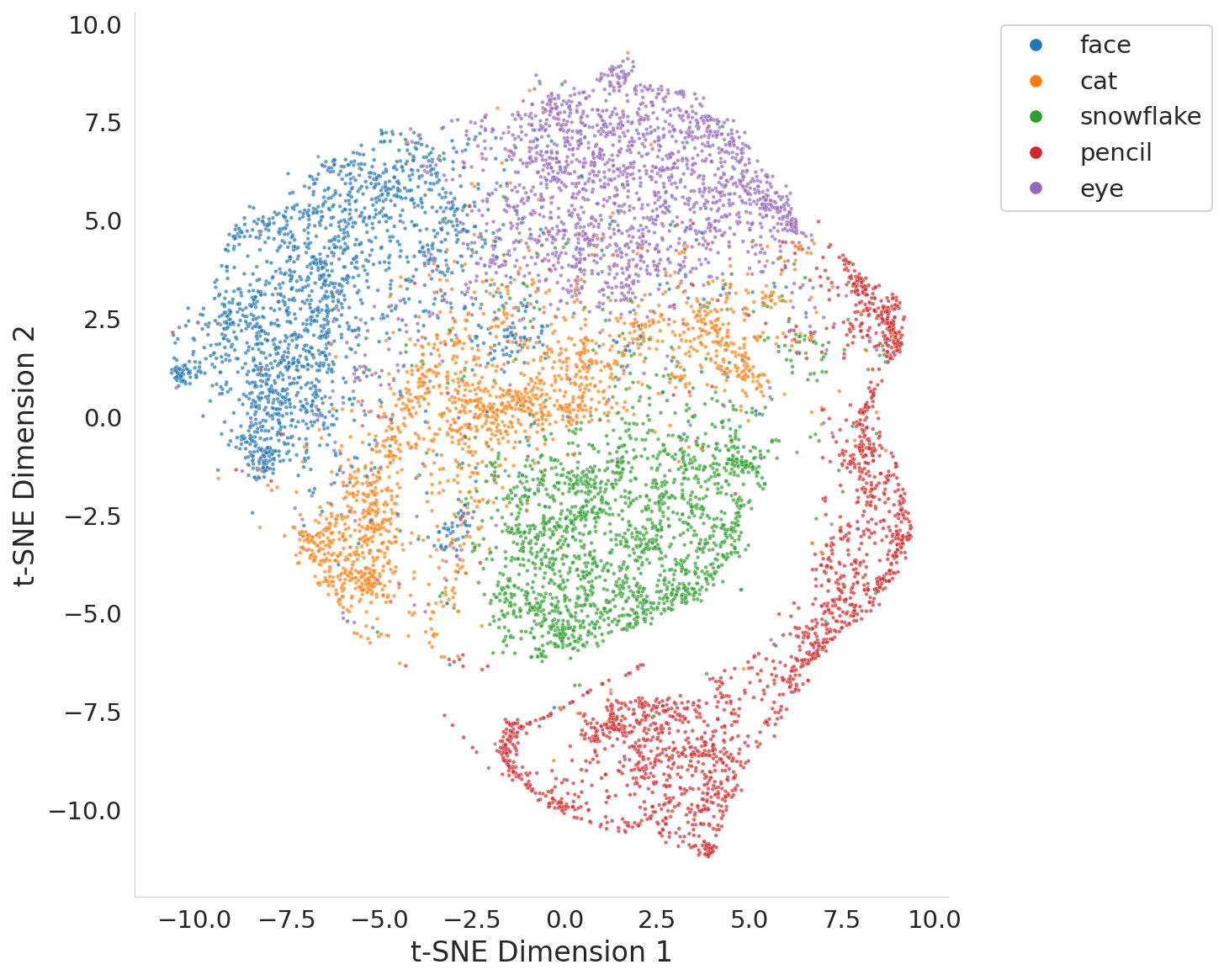}
    \caption{Colored by Class Labels}
\end{subfigure}
\caption{t-SNE of latent space. Expert assignments (left) form more coherent clusters than ground truth class labels (right), explaining the superior performance of unsupervised routing.}
\label{fig:tsne_comparison}
\end{figure}

To understand why unsupervised expert routing outperforms supervised approaches, we analyze how the learned expert assignments relate to the underlying structure of the latent space. Figure~\ref{fig:tsne_comparison} presents t-SNE~\cite{maaten2008visualizing} visualizations of the same latent space representations, colored by expert assignments (left) versus ground truth class labels (right).

The comparison reveals a strong correspondence between the clusters formed by expert assignments and those defined by ground-truth class labels. As seen in Figure~\ref{fig:tsne_comparison}, the spatial organization of expert-colored clusters (left) closely mirrors the class-colored clusters (right). This visual alignment suggests that the unsupervised expert specialization discovers a latent structure that is highly correlated with the semantic categories in the data. To quantify this relationship, we calculated the correlation between the expert assignments through unsupervised training and the class labels, finding a strong positive correlation of 0.802. This indicates that while the model is not explicitly guided by labels, it learns to partition the data in a way that is semantically meaningful.

This correlation between labels and expert assignments is promising, but it does not explain why unsupervised routing performs better than supervised. One visual clue from Figure ~\ref{fig:tsne_comparison} is the better quality of the clustering, which means fewer overlaps between the clusters in case of expert assignments. To quantify this difference in clustering quality, we trained linear classifiers to predict both expert assignments and the original class labels from latent representations. The results strongly support the visual observations: classification accuracy reaches 93.4\% when predicting expert assignments, compared to only 85.1\% when predicting ground truth class labels. This 8.3\% improvement demonstrates that expert assignments are more linearly separable in the latent space.

The experts naturally organize according to the intrinsic geometry of the learned latent space, creating more coherent and well-separated regions that are easier to model with individual decoder networks. In contrast, human-defined categories may not respect the natural boundaries that emerge from the data's underlying manifold structure, leading to expert assignments that are less optimal for reconstruction tasks.

\subsection{Visual Expert Specialization Patterns}

\begin{figure}[htbp]
\centering
\includegraphics[width=0.9\linewidth]{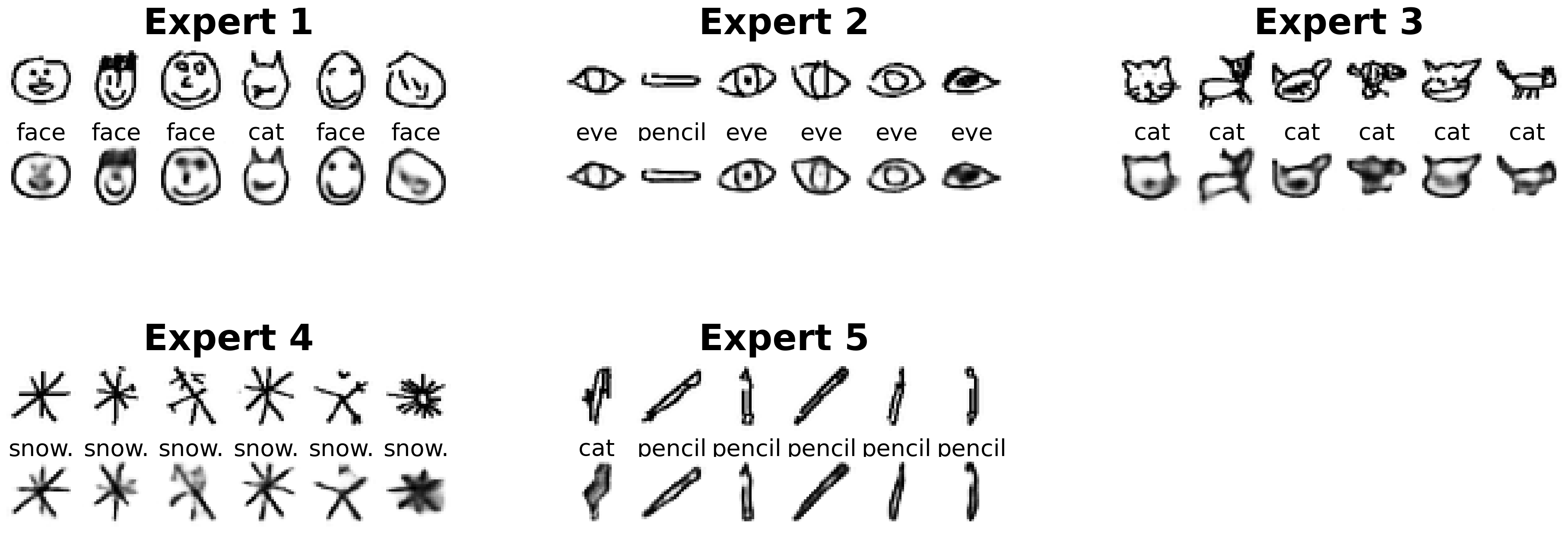}
\caption{Expert specialization for 5 experts. Each expert shows 5 random input images (top) and their reconstructions (bottom), demonstrating specialization based on visual features rather than semantic class labels.}
\label{fig:expert_specialization_5}
\end{figure}

\begin{figure}[htbp]
\centering
\includegraphics[width=\linewidth]{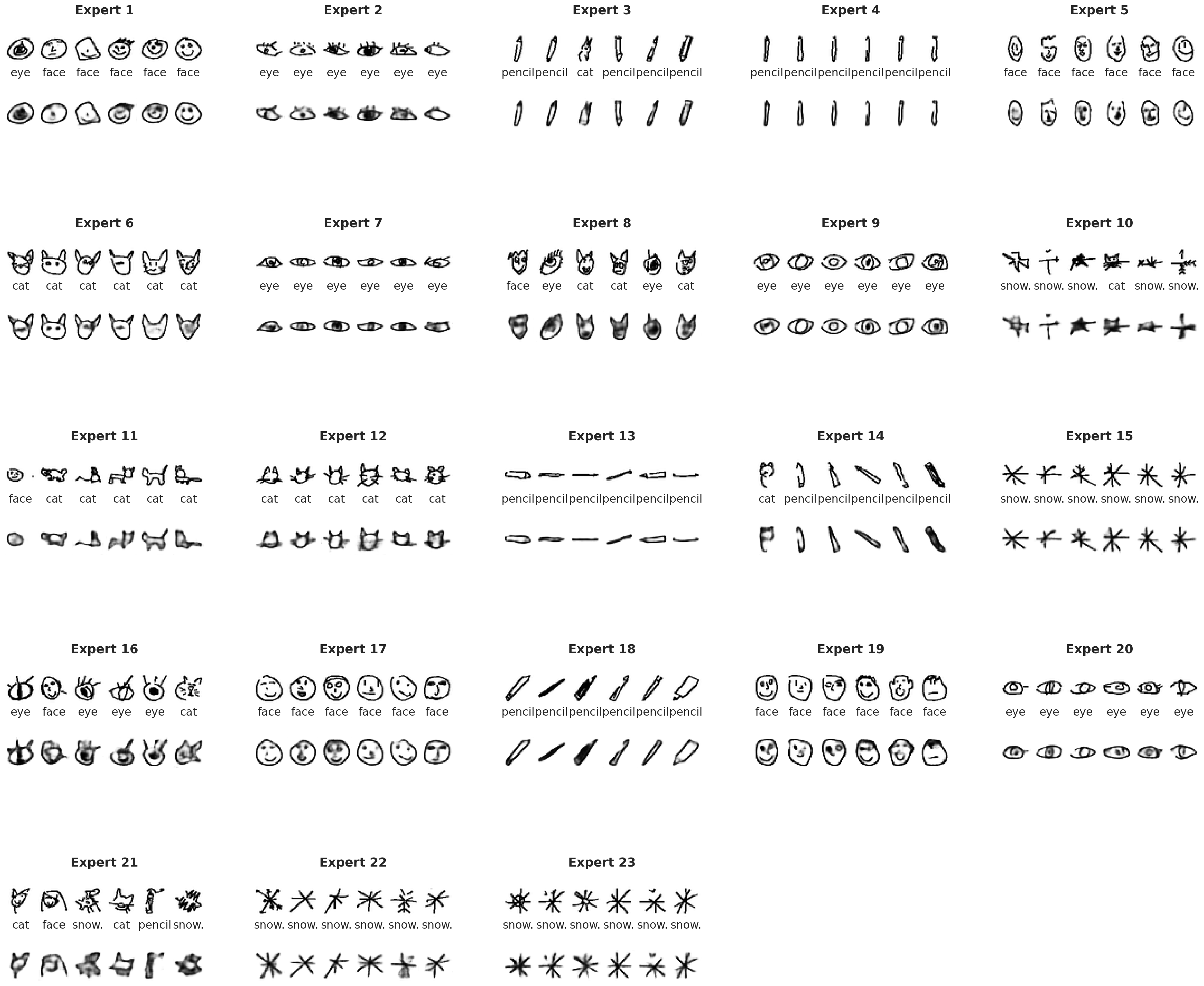}
\caption{Expert specialization for 23 experts showing fine-grained sub-categorical specializations (e.g., pencils at different orientations and various face/eye sub-types).}
\label{fig:expert_specialization_23}
\end{figure}

To directly observe what each expert has learned to specialize on, we visualize the actual images that activate each expert along with their reconstructions. Figure~\ref{fig:expert_specialization_5} shows this analysis for a model with 5 active experts, while Figure~\ref{fig:expert_specialization_23} demonstrates the same for 23 active experts. For each expert, we display 5 randomly selected images that were routed to that expert (top row) and their corresponding reconstructions (bottom row), along with the ground truth class labels and utilization percentages.

The results reveal specialization patterns that transcend class boundaries. In the 5-expert configuration, Expert 1 specializes primarily in faces and cats, Expert 2 focuses on eyes and flat oval structures, Expert 3 handles cats and similar curved shapes, Expert 4 processes snowflakes and Expert 5 is dedicated to pencil-like linear objects. Importantly, the expert assignments capture visual similarity rather than semantic categories, for instance, certain cat drawings that resemble faces are routed to the face expert rather than being constrained by their ground truth labels.

The 23-expert configuration in Figure~\ref{fig:expert_specialization_23} reveals even more granular specialization patterns. Most striking is the fine-grained organization of pencil-like objects: the model learns separate experts for horizontal pencils, vertical pencils, and pencils at various angles. Similarly, cat experts differentiate between different ways of drawing a cat. This level of specialization demonstrates that when given sufficient capacity, the MoE architecture discovers sub-categorical structures that are highly relevant for reconstruction quality but invisible to human categorical labels.

These visualizations provide evidence for why unsupervised expert assignment outperforms supervised approaches. The gating network successfully captures visual similarity patterns that span across traditional class boundaries while discovering meaningful sub-categories within classes. A cat drawing that visually resembles a face is appropriately routed to the face expert, enabling better reconstruction than forcing it into a "cat" expert that may not handle face-like features well. This adaptive specialization based on visual structure rather than assigned labels in part explains the superior reconstruction performance observed in our quantitative results.

The question that arises now is why doesn't the 23 expert configuration outperform the 7 expert one as show in Figure~\ref{fig:supervised_vs_unsupervised}? We conducted further analysis on dataset size impact to answer this question.

\subsection{Dataset Size Impact on Optimal Expert Count}

\begin{figure}[htbp]
\centering
\includegraphics[width=0.9\linewidth]{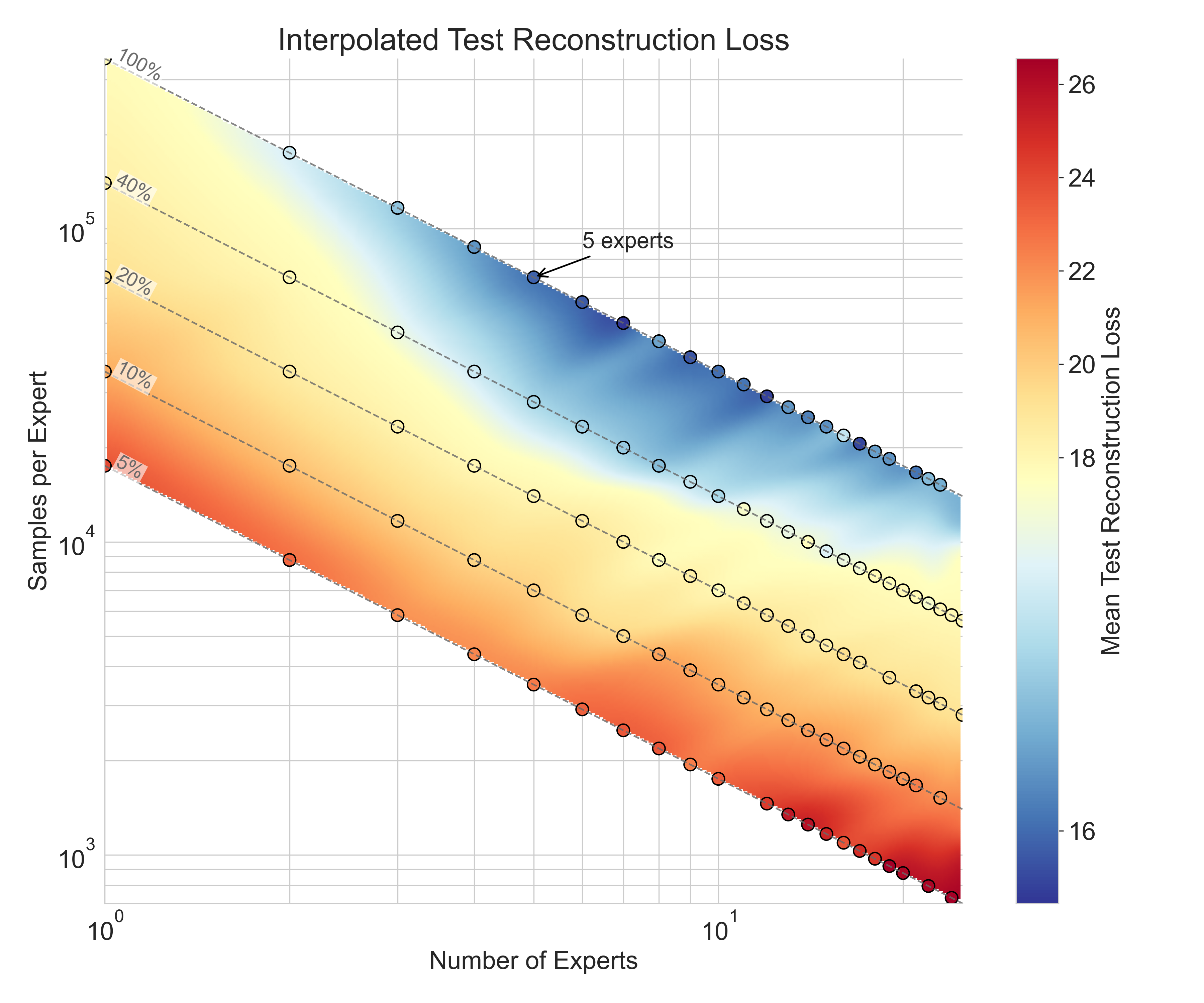}
\caption{Log-log influence of number of experts and samples per expert on performance (test reconstruction loss). Markers show experimental points; surface is linearly interpolated. A clear minimum appears around 5 experts, matching the number of classes.}
\label{fig:dataset_size_impact}
\end{figure}

Previous results raise a question: what factors, beyond the number of semantic categories, influence optimal number of experts? A primary candidate is the size of the dataset, or more specifically, the number of samples available for each expert to learn from.

To investigate this relationship, we conducted a systematic analysis across various data regimes, with the results visualized in Figure~\ref{fig:dataset_size_impact}. This figure presents a log-log plot where the x-axis represents the number of experts and the y-axis represents the number of samples per expert. The color intensity corresponds to the mean test reconstruction loss, with blue indicating lower loss and red indicating higher loss. The surface is generated by linearly interpolating between experimental data points (shown as circles), while the dashed diagonal lines represent fixed percentages of the total dataset size. A complementary view of this data, including error bars, is provided in Appendix \ref{sec:appendix_a}.

The plot reveals several key trends. A distinct valley of low reconstruction loss is centered near 5 experts, which aligns with the number of classes in our dataset. This suggests that, as a baseline, the model's performance is optimized when the number of experts roughly matches the number of primary data categories. 

If we observe the plot vertically (fixing the number of experts and increasing samples per expert) we confirm a well-known principle: more data leads to better performance, as evidenced by the color shifting from red to blue. A more interesting pattern emerges when we analyze the plot horizontally, increasing the number of experts while keeping the number of samples per expert constant. If the number of samples per expert was the sole factor that influenced expert performance (lower loss) we would expect to observe somewhat constant performance along the horizontal axis. Instead, we observe a significant improvement in performance as we move to the right. This phenomenon can be attributed to two factors. First, although each decoder expert is trained on a fixed number of samples, the shared encoder is exposed to a larger and more diverse dataset, allowing it to learn more robust and effective latent representations for the entire network.

Second, and more fundamentally, increasing the number of experts allows for greater specialization. With fewer experts, each is forced to model a more complex, multi-modal data distribution (e.g., a single expert may have to learn to reconstruct both "cat" and "pencil" sketches). In contrast, with more experts, each can specialize on a simpler, more uni-modal subset of the data. To validate this hypothesis, we conducted an additional experiment: a single expert trained on the five-class dataset achieved a reconstruction loss of 24.0. When the same network was trained on single-class datasets of the same total size, average loss was significantly lower, at 18.4. For comparison, the same network trained on 20 times more data achieves loss of 17.7, which is only marginally better than the previous. This confirms that it is more effective for experts to learn from homogeneous data distributions, explaining why the performance improves as the model is given more experts to partition the data, even though the number of samples per expert remains constant.

\section{Conclusion}

We have presented a novel SMoE-VAE architecture that enables interpretable analysis of expert specialization through visual reconstruction patterns. Our key finding demonstrates that for the QuickDraw database, unsupervised expert routing consistently outperforms supervised routing based on human-provided labels, revealing that experts naturally discover data structures which are more informative, when allowed to specialize according to intrinsic patterns rather than categorical boundaries.

Through comprehensive evaluation on the QuickDraw dataset, we showed that experts develop coherent specializations that span across traditional class boundaries while identifying reconstruction-relevant subclusters within classes. The t-SNE analysis aand visual reconstruction patterns provide compelling evidence that expert assignments based on learned data structure achieve superior linear separability (93.4\% vs 85.1\%) and reconstruction quality compared to human-defined categorizations.

Our analysis of dataset size effects reveals a nuanced relationship between performance, data quantity, and the degree of expert specialization. We found that expert performance is more sensitive to the homogeneity of the data it models than to the absolute number of samples it is trained on. Increasing the number of experts allows for greater specialization on simpler, more uni-modal data subsets, which improves reconstruction quality even when the number of samples per expert is held constant. However, this benefit is balanced by the risk of data starvation; for a fixed dataset size, increasing the number of experts indefinitely degrades performance. These findings highlight a critical trade-off in MoE design and provide guidance for selecting an appropriate number of experts based on the underlying structure and size of the dataset.

At this point, it is worth mentioning that our experiments had a persistent challenge with expert collapse, where roughly half of the available experts remain inactive during training. This suggests potential issues with network initialization or the early training procedure that warrant further investigation. 

Another key point is that our load balancing and entropy regularization terms differ from standard approaches in the literature, with ~\cite{zhang2017moe} using almost the same loss formulation. However, we note that the theoretical work of~\cite{understanding_moe} achieved low entropy probability distributions with equal load balance, providing strong indications that our conclusions would remain valid with alternative loss formulations.

Looking toward future research, several important directions emerge from this work. First, synthetic datasets could provide controlled environments for a deeper analysis of expert specialization mechanisms. Second, the behavior of SMoE-VAE architectures under dataset imbalance conditions remains unexplored and could reveal important robustness characteristics. Third, extending our approach to multiple layers of MoE could unlock more sophisticated hierarchical specialization patterns. Lastly, one obvious extension of our work is to validate the findings across different image databases.

Our work sugests that expert specialization can serve as a lens for understanding fundamental data organization principles. The finding that unsupervised routing discovers a more informative structure than human categorizations has broader implications for how we conceptualize optimal computational organization in neural networks. This methodology opens new avenues for interpretable analysis of complex architectures and provides a foundation for future investigations into the relationship between learned representations and labeled data structure.

\appendix
\section{Different look on Figure~\ref{fig:dataset_size_impact}}
\label{sec:appendix_a}

\begin{figure}[htbp]
\centering
\includegraphics[width=0.9\linewidth]{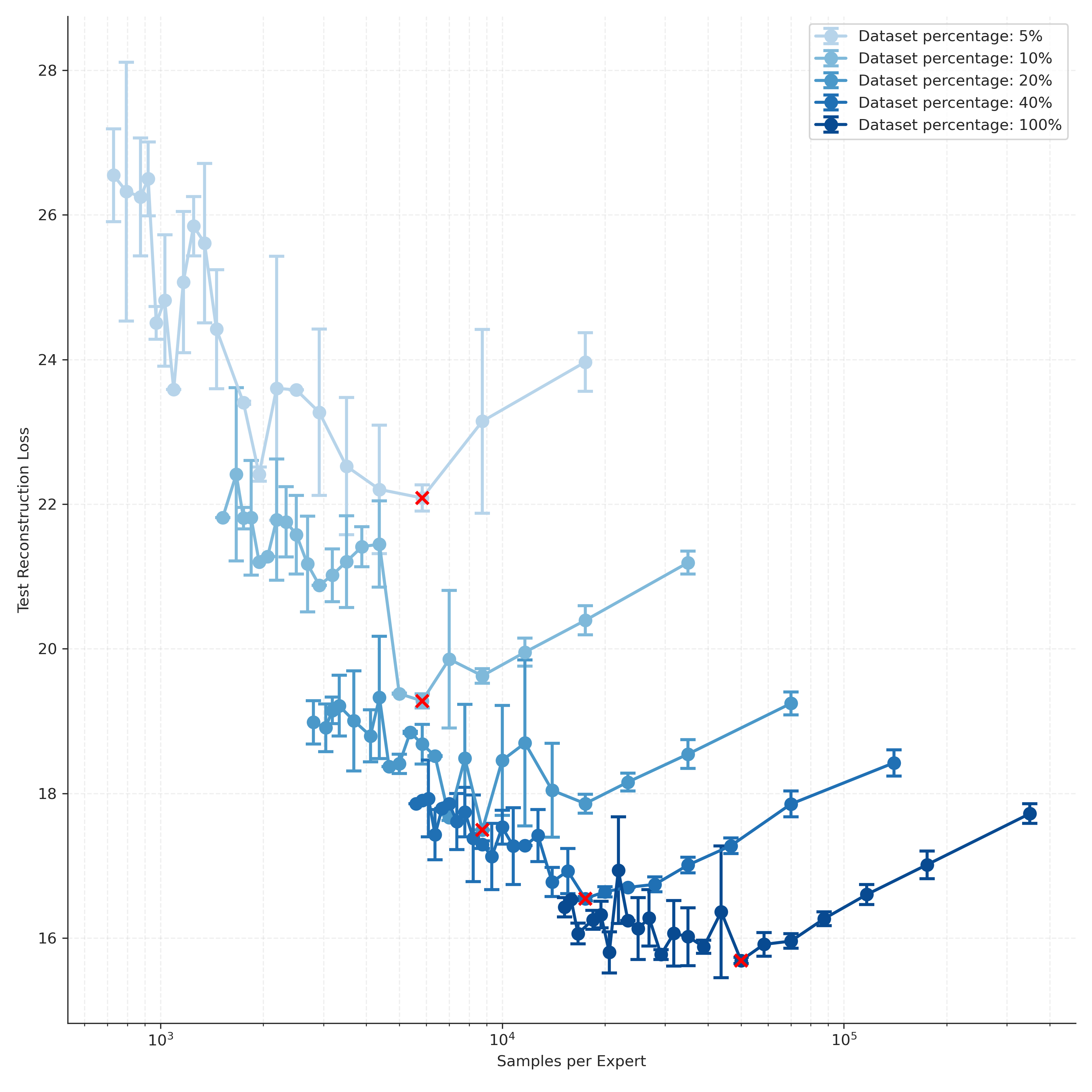}
\caption{Test reconstruction loss as a function of samples per expert for different dataset sizes. Each curve represents a fixed dataset percentage, with error bars showing standard deviation across runs.}
\label{fig:dataset_size_impact_appendix}
\end{figure}

Figure~\ref{fig:dataset_size_impact_appendix} provides a complementary view to Figure~\ref{fig:dataset_size_impact}, making it easier to see differences in test reconstruction loss in absolute numbers, with explicit error bars indicating the standard deviation across multiple runs. Each curve corresponds to a different percentage of the total dataset, from 5\% to 100\%. The x-axis represents the number of samples available to each expert, which means that the number of active experts increases from right to left along each curve. Note that the lowest curve (with 100\% of data used) uses the same data as the curve in Figure~\ref{fig:supervised_vs_unsupervised}.

This visualization offers another perspective on the relative importance of data homogeneity versus the sheer volume of samples per expert. For instance, a model trained on 5\% of the dataset (the light blue curve) with 17,500 samples allocated to a single expert achieves a reconstruction loss of around 24.0. In contrast, a model trained on 40\% of the data (the dark blue curve) but with a same number of samples per expert distributed among 4 experts achieves a significantly lower loss of approximately 18.0. This comparison reinforces the conclusion from the main text: expert performance is more sensitive to the homogeneity of the data it models than to the absolute number of samples it is trained on.

Furthermore, this plot helps explain why increasing the number of experts does not always lead to better performance. As the number of experts grows (moving from right to left on a given curve), the number of samples per expert decreases, eventually leading to data starvation and degraded performance.

\section*{Code and data availability}
Our code is available at \url{https://github.com/strajdzsha/smoe-vae}.

\end{document}